\def\year{2022}\relax
\documentclass[letterpaper]{article} % DO NOT CHANGE THIS
\usepackage{aaai22}  % DO NOT CHANGE THIS
\usepackage{times}  % DO NOT CHANGE THIS
\usepackage{helvet}  % DO NOT CHANGE THIS
\usepackage{courier}  % DO NOT CHANGE THIS
\usepackage[hyphens]{url}  % DO NOT CHANGE THIS
\usepackage{graphicx} % DO NOT CHANGE THIS
\usepackage{natbib}  % DO NOT CHANGE THIS AND DO NOT ADD ANY OPTIONS TO IT
\usepackage{caption} % DO NOT CHANGE THIS AND DO NOT ADD ANY OPTIONS TO IT
\DeclareCaptionStyle{ruled}{labelfont=normalfont,labelsep=colon,strut=off} % DO NOT CHANGE THIS
\usepackage{algorithm}
\usepackage{algorithmic}
\usepackage{newfloat}
\usepackage{listings}
\floatstyle{ruled}
\newfloat{listing}{tb}{lst}{}
\floatname{listing}{Listing}
\usepackage{amssymb}
\usepackage{booktabs}
\usepackage{multirow}
\usepackage{tabularx}
\usepackage{amsmath}
\usepackage{xspace}

\newcommand{\eg}{\emph{e.g.,}\xspace}

\newcommand{\ie}{\emph{i.e.,}\xspace}

\newcommand{\etal}{\emph{et al.}\xspace}

\title{Gradient-guided Unsupervised Text Style Transfer via Contrastive Learning}
\author{
    Chenghao Fan\textsuperscript{\rm 1,\footnote{Equal contribution.}},
    Ziao Li\textsuperscript{\rm 1,*},
    Wei wei\textsuperscript{\rm 1,\footnote{Corresponding authors.}}
}
\affiliations{
    \textsuperscript{\rm 1}School of Computer Science, Huazhong University of Science and Technology, Hu’bei, China\\
    \{facicofan,leeziao,weiw\}@hust.edu.cn\\
}
\usepackage{bibentry}
\begin{document}

\maketitle

\begin{abstract}
Text style transfer is a challenging text generation problem, which aims at altering the style of a  given sentence to a target one while keeping its  content unchanged.  Since there is a natural scarcity of parallel datasets, recent works mainly focus on solving the problem in an unsupervised manner. However, previous gradient-based works generally suffer from the deficiencies as follows, namely: (1) Content migration. Previous approaches lack explicit modeling of content invariance and are thus susceptible to content shift between the original sentence and the transferred one. (2) Style misclassification. A natural drawback of the gradient-guided approaches is that the inference process is homogeneous with a line of adversarial attack, making latent optimization easily becomes an attack to the classifier due to misclassification. This leads to difficulties in achieving high transfer accuracy. To address the problems, we propose a novel gradient-guided model through a contrastive paradigm for text style transfer, to explicitly gather similar semantic sentences, and to design a siamese-structure based style classifier for alleviating such two issues, respectively.  Experiments on two datasets show the effectiveness of our proposed approach, as compared to the state-of-the-arts.
\end{abstract}
%In attempt to address the above issues, we propose our gradient-guided model, which explicitly cluster content-similar sentences and utilize a siamese structure style classifier to tackle the above issues respectively.
%To the best of our knowledge, we are the first to use contrastive learning in the domain of text style transfer. Furthermore, our model has the ability to do a fine-grained transfer, where the intensity of the target style can be controlled. Experiments on two datasets show the effectiveness of our approaches on resolving the proposed issues.
%Our model achieves state-of-the-art performance on both automatic and human evaluations. Our code and data are availble at http://I don't know where is my code and data.
%Gradient-based update has been widely adopted in the regime of conditional generation, including text style transfer.
%Previous gradient-based methods, including the text style transfer ones, 
%suffers from the following drawbacks: Unconstrained Nature Of Content (UNOC) 
%and Unstability Of Embedding Classifier (UOEC).

\section{Introduction}

Text style transfer, as an important task of natural language generation (NLG), aims at altering the style of a given sentence (\eg positive) to a target one (\eg negative) while preserving its content as much as possible.
The controllable rewriting a sentence with desired style is beneficial for many downstream applications in practice, such as converting offensive language to non-offensive \cite{tran2020towards}, converting biased remarks to neutral \cite{pryzant2020automatically} and generating eye-catching headlines \cite{headline1,headline2}.
Moreover, text style transfer may serve as data augmentation for many natural language subtasks, and thus it has attracted a considerable amount of research.

Since there is a natural scarcity of parallel datasets, recent works mainly focus on solving the problem in an unsupervised manner, where only labelled sentences are available. 
% Many attempts have been made to address this task.
Several efforts have been devoted on the gradient-guided optimization based models, \eg \cite{19nips,20aaai}, which are usually consist of two subcomponents: (1) Auto-Encoder, which learns the mapping between a discrete sentence space and a continuous latent space; (2) Style classifier, which predicts the style type of a decoded sentence based on its latent representation. The representation of the original sentence is edited iteratively to the target one during inference, along with the direction of the gradient obtained from the style classifier.
% One line of works is based on gradient-guided optimization, such as \cite{19nips,20aaai}.
% The typical workflow of such methods is: 
% (1) Train an auto-encoder which learns a mapping between the discrete sentence space and the continuous latent space.
% (2) Train a style classifier which takes the latent representation and predicts the style of its corresponding decoded-sentence. At inference stage, latent embeddings of original sentences are edited iteratively in the direction of their gradients obtained from the style classifier to get target embeddings which belong to the desired style.

Nevertheless, previous gradient-guided approaches generally suffer from the deficiencies as follows: 
(1) Content migration. Content invariance is of crucial importance to evaluate the success of a text style transfer model, however, nearly none of existing works takes account of an explicit constraint to ensure the content invariance before and after conversion, which may result in a great discrepancy of the content between the original sentence and the target one.
Later, there are various attempts on content consistency, for example, Liu \etal \shortcite{20aaai} propose a content predictor to tackle such problem by predicting the word features (\ie Bag-of-Words) of the generated sentence during inference.
Nevertheless, experiments illustrate that such method result in trivial improvement. 
% (2)
% (1) the under-constraint of preserving content. 
% Content preservation is crucial to a successful text style transfer, however, existing gradient-guided works lacks explicit constraint to force sentence content before and after transferring related, and thus produce a large content variation between original sentences and transferred ones. 
% Though \cite{20aaai} attempts to gain better content preservation by introducing a content predictor, which takes the latent representation and predict the Bag-of-Word feature of its decoded sentence. 
% (2) the inaccuracy of style classifier. 
(2) Style misclassification.
A robust style classifier is vital in gradient-guided methods, as it provides the direction for the refinement of latent representation during inference.
% since the direction of embeddings optimization is obtained from it. 
However, the process of searching target embedding through gradient optimization resembles a line of attacking white-box neural network, 
for example, Hsieh \etal \shortcite{19attack} attack the style classifier by applying gradient-based perturbations.
As a result, an expected style transformation may become an attack to the style classifier due to misclassification, which brings about difficulties in achieving high transfer accuracy.
% Our experiments further manifest a phenomenon that two embeddings are decoded to be the same sentence but are classified to be different styles.
% In consequece, previous works have difficulties in achieving high transfer accuracy.

To address these problems, we propose a novel gradient-guided model for text style transfer.
We adopt a contrastive paradigm to train a better auto-encoder and design a more robust siamese-structure based style classifier for alleviating such two issues, respectively. 
With respect to the first issue, it is worth noting that transferred sentences are adjacent to the original ones in terms of embedding distance, since the gradient update steps are minimal. 
Therefore, sentences is capable of being optimized to the desired ones only if they are neighboring in the latent space. 
Accordingly, we adopt a contrastive paradigm for training the auto-encoder, which explicitly models content invariance by drawing embeddings of similar content sentences closer and pushing those of different content apart. 
With respect to the second issue, we design a novel siamese-structure based style classifier.
The classifier takes two sentences as input and yields their likelihood of being the same style.
In such wise, our classifier predicts the style of an embedding by conducting comparison with other labelled sentence embeddings, and thus the accuracy of style identification increases when the number of labelled references increments. 
Experiments show our proposed siamese-structure based style classifier is more resistant to style misclassification.
% and improves transfer accuracy in the task of text style transfer. 
% The adoption of our siamese-structure classifier improves style accuracy in the task of text style transfer.

Our contributions are summarized as follows: 

\begin{itemize}
  \item We analyze the cause of content migration of gradient-guided approaches and correspondingly propose an auto-encoder with a contrastive paradigm, which effectively improves content consistency before and after conversion. 
  \item We design a novel siamese-structure based classifier, which is more resistant to style misclassification and improves style transfer accuracy.
  \item Experiments shows our model achieves state-of-the-art performance in both automatic and human evaluation.
\end{itemize}

\section{Background}
\subsection{Style Transfer}

Style transfer is a task targeting at changing the stylistic attribute while retaining the content of the input text. 
Owing to the lack of parallel corpora, recent methods mainly work in an unsupervised manner. 
Most of previous approaches address the task with a latent manipulation workflow: first encode original sentences into latent representations; then manipulate the latents; finally feed the latents to a decoder to generate target sentences. 
\cite{crossalign} assumes a shared latent content distribution across the corpus of different styles. 
\cite{towardscontrolled} utilizes the wake-sleep algorithm for learning a structured style code. 
\cite{john2019disentangled} designs multiple adversarial losses to achieve a separation of style and content latent representations. 
% \cite{bt1} adopts back-translation technique in attempt to import more prior knowledge. 
% \cite{towardscontrolled,crossalign,bt1,john2019disentangled} explicitly disentagle the latent representation of sentences into style and content representations. At inference stage, orginal style representations is replaced with target ones while the content representations unchanged to conduct style transfer.
\cite{delete} adds an embedding with target style to the entire representation instead of separating style and content.
\cite{styins} constructs a style space to sample more diverse style embeddings.
\cite{huang2019dictionary} implements an attention mechanism to achieve phrase level style representations. 
\cite{pryzant2020automatically} introduces a tagger module which adds style embeddings of different intensity to different words in a sentence. 

% Researches which do not require manipulation of latent representations have also been conducted. \cite{imat} build pseudo-parallel corpus to overcome the shortage of scarcity in parallel datasets. \cite{hierarchical,miao2019cgmh} identify attribute words and select another one to replace it.

% Our line of gradient-guided text style transfer \cite{19attack,20aaai} requires manipulation of latent representaions.

% Among these approaches, there is a line of works aims to generate a latent representation for the input sentence, and
% change the style of the generated sentence based on the learned latent representation.
% For this strategy, most solutions use adversarial methods \cite{bt1, towardscontrolled,crossalign}
% to learn the representation of the style and content, then combining the style and 
% content latent representations or use specific decoders to generate different 
% attribute-specific sentences. However, some experiments show that disentanglement is 
% unnecessary \cite{lample2018multiple,19nips,20aaai}. These works show that a good decoder can generate the 
% text with the desired style from an entangled latent representation by “overwriting” 
% the original style. Different from their approaches, we propose an entangled style translation model which uses 
% contrastive learning to construct the autoencoder. 

\subsection{Contrastive Learning}

Contrastive learning is proved to be an effective unsupervised method for learning an expressive latent representation space \cite{chen2020big,he2020momentum} through pulling semantic similar neighbors closer and pushing non-neighbors apart \cite{hadsell2006dimensionality}. 
Recently, the contrastive manner has shown effectiveness in learning better representations both in CV \cite{chen2020big,he2020momentum} and NLP \cite{kaushik2019learning,gao2021simcse,carlsson2020semantic}. 
% Many approches have been studied for finding similar neighbors, for example crop and rotate a picture in the regime of CV and delete and replace a word in the regiem of NLP.

% Contrastive learning aims to learn effective representation by pulling semantic similar neighbors closer and pushing non-neighbors apart \cite{hadsell2006dimensionality}.
% Recently, constrastive learning has become a popular approach in the field of unsupervised visual representation learning, and has a good performance \cite{chen2020big, he2020momentum}.

% They apply image transformation to randomly generate two augmented samples from each image and make them close in representation space.
% Similar to the CV field, contrastive learning has recently performed well in NLP.
% The contrastive manner has shown effectiveness in learning sentence representation\cite{kaushik2019learning,gao2021simcse,carlsson2020semantic}.

Our model follows the contrastive framework in \cite{chen2020big} and applies a normalized temperature-scaled cross-entropy loss with in-batch negatives in our Transformer-based auto-encoder. 
We assume a set of paired examples $D=\{(x_i,x_i^+)\}_{i=1}^n$, where $x_i$ and $x_i^+$ are content-similar. 
For a batch with $N$ pairs, the training objective for $(x_i,x_i^+)$ is: 
\begin{equation}
con_i = -\log {\frac{e^{sim(r_i,r_i^+)/\tau}}{\sum_{j=1}^N e^{sim(r_i,r_j^+)/\tau}}}, \label{eq:1}
\end{equation}
where $\tau$ is a temperature hyperparameter, $sim(\cdot)$ indicates cosine similarity function, 
$r_i$ and $r_i^+$ denote the encoded representations of $x_i$ and $x_i^+$.
% Through the contrastive learning, we draw content-similar sentences closer in the latent space so that the generated sentence by Gradient-guided Update preserves more content than previous lines.

\subsection{Gradient-guided Optimization}

Gradient-guided methods have been widely used for controllable generation, which edits the latent representations according to its gradient obtained from a neural network.
Such methods require two subcomponents: an auto-encoder which learns a mapping between the source data distribution and continuous latent space and a neural network which is trained to discriminate task-specific features, \eg a style classifier for recognizing input styles.
At inference stage, we generate the target with following three steps: 
(1) Encode the source (\eg sentence) into a continuous latent representation with a pre-trained auto-encoder. 
(2) Edit the latent representation according to its gradient, which is obtained by back propagating the task-specific neural network.
(3) Decode the altered latent representation and get the transferred target with the auto-encoder. 

% \begin{enumerate}
%     \item Encode the source (\eg sentence) into a continuous latent representation with a pre-trained auto-encoder. 
%     \item Edit the latent representation according to its gradient, which is obtained by back propagating the task-specific neural network.
%     \item Decode the altered latent representation and get the transferred target with the auto-encoder. 
% \end{enumerate}

% 1) Encode the text into a continuous latent space with a pre-trained autoencoder. 
% 2) Change latent representations according to its gradient. 
% 3) Decode the latent representations to target transferred text based on the pre-trained autoencoder. 
The most universal line to acquire the gradient is to train a classifier which outputs the probability of each class directly \cite{cvpr2017plug,19nips,20aaai}. 
The classifier is trained by minimizing: 
% $$\mathcal{L}_{C}=\mathbb{E}_{z\sim q_{E}(z|x,\theta_{enc})}[p(\hat a_i |a_i)],$$k
\begin{equation}
    \mathcal{L}_{Cls}=-\mathbb{E}_{z\sim q_{E}(z|x,\theta_{enc})}\sum_{i=1}^k\hat{a_i} \log[p(a_i |z)]
\end{equation}
% Such methods optimize the following loss function at generation time:
where $a$ denotes the predicted probability distribution of style $s$ and $\hat a_i$ denotes the true probability distribution.

At inference time, the parameter of classifier is frozen and latent representations are optimized in the direction of minimizing $\mathcal{L}_{Cls}$: 
\begin{equation}
\hat z=z-\omega\cdot\nabla_z\mathcal{L}_{Cls} 
\end{equation}
where $\hat a$ in the inference process is the desired style distribution for the source to transfer.

% To implement style transfer, for example transferring into style $s_i$, the true probability distribution $\hat{a}$ should be $1$ only at $a_i$ and otherwise $0$.

% Since the classifier is based on the embedding which is entangled with content and attribute,
% the classifier may focus on differences in some style-independent features to achieve the 
% effect of binary classification. This may result in changing sentences with only a few
% differences in content rather than focusing on changing some stylistic features.
% So we use the contrastive learning to clusering the texts with similar content in latent space in order
% to easily transfer texts to different styles with similar content.(need experiments)

% Additionally, in the case of using a normal classifier, it may happen that a few small perturbations are made
% to fool the classifier when changing the gradient. So we propose a siamese structure to replace the normal classifier.
% We assumes a set of paired examples $D=\{(x_i,{x_i^0}^+,...,{x_i^n}^+, {x_i^0}^-,...,{x_i^m}^-)\}$,
% where $x_i$ and ${x_i^j}^+$ have a common style,  $x_i$ and ${x_i^j}^-$ have a different style. 
% For a batch with $N$ pairs, the training objective for $(x_i,{x_i^0}^+,...,{x_i^n}^+, {x_i^0}^-,...,{x_i^m}^-)$ is:

% $$l_i=\sum_{k=1}^n -\log {\frac{exp(sim(r_i,{r_i^k}^+)/t)}{exp(sim(r_i,{r_i^k}^+)/t) + \sum_{j=1}^m exp(sim(r_i,{r_i^j}^-)/t)}}, \label{eq:2}$$
% We can also set some hard negative examples so that make our classifier more robust and stable, which can be 
% demonstrated by the experiments that follow

\section{Our Method}

\subsection{Problem Formalization}

The unsupervised text style transfer task can be formalized as follows: 
let $D$ be the dataset, which contains $n$ labelled sentences, namely, 
$D=\{(x_i,s_i)\}_{n}$, 
where $x_{i}$ denotes the text, $s_i \in S$ the corresponding style label and $S$ the set of all styles (\eg $S=\text{\{"positive", "negative"\}}$ for style transfer task). 
% Each $s$ has $k$ attributes of interest $s=\{s_1,...,s_k\}$.
The goal of style transfer problem is to take $(x,s_{tgt})$ as input and output sentence $\hat{x}$ with style $s_{tgt}(s_{src} \neq s_{tgt})$, where $x$ is a sentence with style $s_{src}$. 

% given a sentence $x$ with a style label $s_{src}$ and a desired style $x_{tgt}$, convert it to a new one $\hat x$ with a target style $s_{tgt}(s_{src} \neq s_{tgt})$ while preserving its original content.

\subsection{Model Overview}

\begin{figure}[t]
    \centering
    \includegraphics[width=0.95\columnwidth]{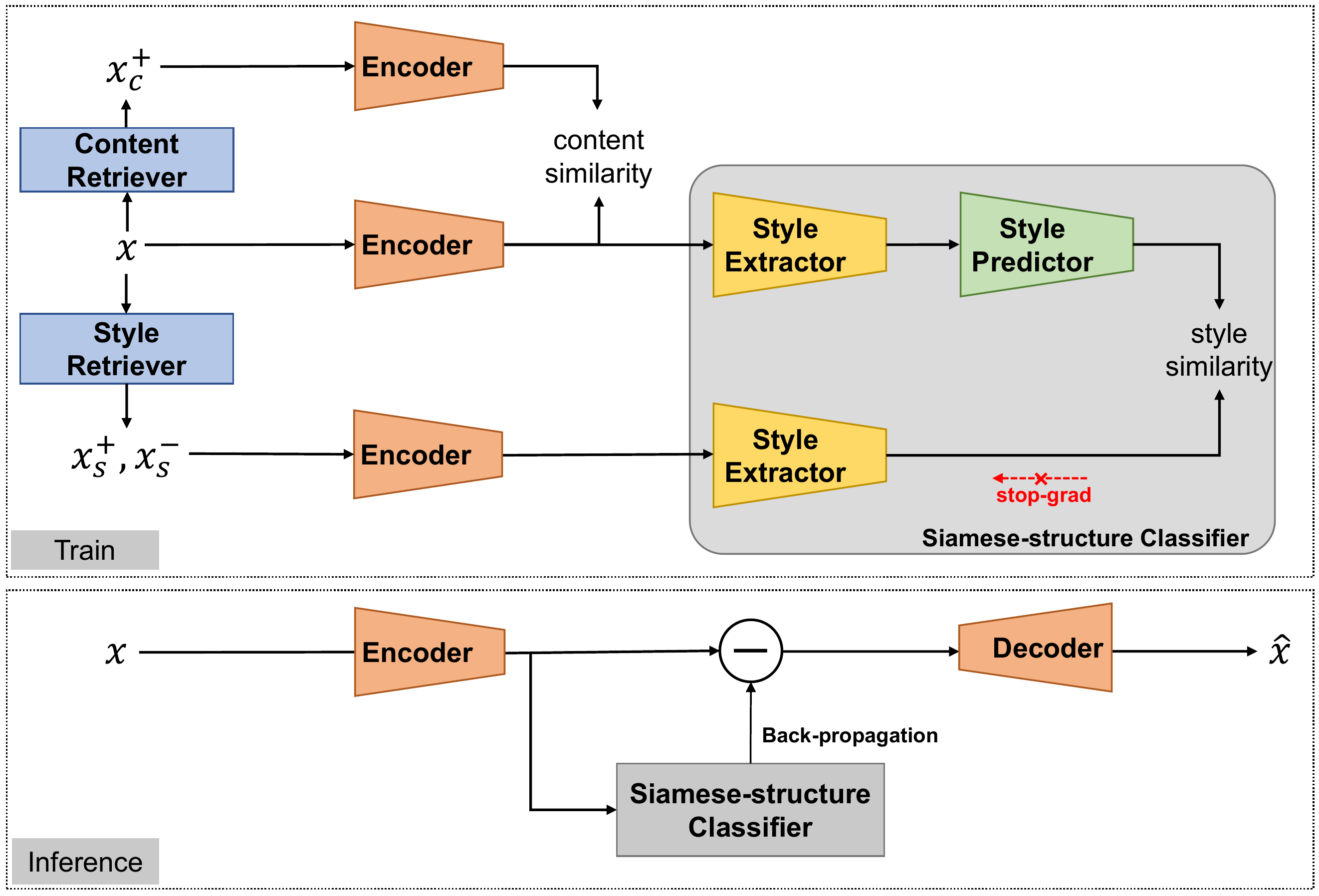}
    \caption{The architecture of our model. The top plot indicates the training stage and the bottom plot the inference stage. The content similarity generated by Contrastive Learning 
    which is used for constraint to contents. The style similarity generated by our Siamese Structure Classifier which is used for predicting the style.}
    \label{fig:attack}
\end{figure}

The proposed model consists of two components: an auto-encoder which adopts a contrastive paradigm to learn a mapping function between texts and latent representations, a siamese-structure based style classifier which identifies style differences between embeddings and provides gradient-guided information for manipulating latent representations. 
The training and inference procedure is shown in top plot and bottom plot of Fig.\ref{fig:attack}, respectively.

\subsection{Auto-encoder}

% Due to the good performance of Transformer \cite{transformer} in the field of text generation, 
We follow the standard encoder-decoder architecture to build a Transformer-based \cite{transformer} one with 
reconstruction loss $\mathcal{L}_{rec}$. 
% Meanwhile, we draw content-similar sentences closer by contrastive learning, which can
% be optimized to minimize the loss $\mathcal{L}_{CL}$.
Explicitly, given a sentence $x$, the Transformer encoder $Enc(x;\theta_{enc})$
maps $x$ to relevant continuous representation $z$ which is entangled with content and style 
and the Transformer decoder $Dec(z;\theta_{dec})$ maps latent representation $z$ back to the sentence $x$. 
Suppose the latent representation $z$ follows the distribution $q_{E}(z|x,\theta_{enc})$ and the sentence $x$ follows the distribution $p_G(x|z,\theta_{dec})$, 
our auto-encoder reconstruction loss is formalized as: 
\begin{equation}
 \mathcal{L}_{rec}=-\mathbb{E}_{q_{E}(z|x,\theta_{enc})}[\log p_G(x|z,\theta_{dec})]   
\end{equation}

\subsection{Modeling Content Invariance}

To add explicit constraint on content consistency in the standard Transformer framework, 
we adopt a contrastive paradigm for training a better auto-encoder, which models content invariance in entangled latent space. 

%Therefore to perform a style transfer task, the target sentence must be very adjacent 
%to the source one in the representation space. In addition, sentences with similar 
%content should cluster in the representation space, which can avoid the lost of 
%original content and the export of other unintentional content.
%To deal with above two hypothesis, we propose a contrastive loss when to train the auto-encoder, 
%for the purpose of getting a better representation.

Explicitly, the contrastive paradigm is composed of the following two parts:  
(1) draw the sentences closer, which are content-similar but have different style. 
(2) draw the sentences closer, which are content-similar and have same style. 
By this means, sentences with similar content cluster together in the latent space. 
The motivation is that transferred embeddings are adjacent to original ones, since gradient-guided optimization steps are minimal. 
Therefore, a sentence can be optimized from the original embedding only if they are close to each other in terms of latent representations distance. 
Accordingly, the approach of drawing style-different content-similar sentences closer makes the gradient-optimization process easier to implement, and the approach of drawing style-similar content-similar sentences closer prevents occurrence of original content lost and export of other unintentional content. 
% which avoids the lost of original content and the export of other unintentional content.
Nonetheless, A challenge is that we have no access to the required content-similar paired corpus, neither of the same style nor different styles, for contrastive learning.
To solve the problem, we use retrieval and data augmentation to construct pseudo data pairs. 
For the purpose of better illustration, we exemplify our model with sentence $x_i$ whose style is $s_i$. 

For the first part,
% suppose our target is to change the sentence style from $s_i$ to $s_j$,
we measure the semantic similarity of two sentences by Sentence-BERT (SBERT) \cite{reimers2019sentence} which
is capable of efficiently computing semantic textual similarity.
For the original sentence $x_i$, we construct $(x_i,x_i^{like})$ as a pseudo parallel data pair, where $x_i^{like}$ is the sentence of style $s_j(s_j \neq s_i)$ that shares most content similarity with $x_i$.
To ensure that the retrieved sentences pairs have similar semantics, we set a SBERT threshold $\beta$ for retrieving sentences pairs, and only pairs above this threshold are used for constructing our pseudo data pairs.
In addition, to speed up the search in a large corpus, we retrieve the most similar sentence for each sample with the help of FAISS\footnote{https://github.com/facebookresearch/faiss}.
Borrowing the supervised contrastive loss from \cite{supervised_contrastive}, we use these data pairs to optimize Eq.\ref{eq:1} and then the contrastive loss is formulated as follows:
% we use them to optimize
% \ref{eq:1} and get the loss $\mathcal{L}_{CL_{diff}}$:
\begin{equation}
   {cons_{diff}}_i = -\log {\frac{e^{sim(z_i,z_i^{like})/\tau}}{\sum_{j=1}^N e^{sim(z_i,z_j^{like})/\tau}}} 
\end{equation}
\begin{equation}
    \mathcal{L}_{CL_{diff}}=\sum_i {cons_{diff}}_i
\end{equation}
where $z_i$ and $z_i^{like}$ denote the representations of $x_i$ and $x_i^{like}$.

For the second part, 
we construct pseudo data pairs $(x_i,x_i^{drop})$ through adding dropout perturbations which is demonstrated to be able to learn a good alignment for positive pairs \cite{gao2021simcse}. 
The retrieved sentence $x_i^{drop}$ is the same as original sentence $x_i$, however their representations $z_i$ and $z_i^{drop}$ differ due to the existence of random dropout when encoding a sentence. 
% We define the pairs we get by this method is the form of $(x_i,x_i^{+})$. 
Likewise, we use them to optimize Eq.\ref{eq:1} like part one and get the loss $\mathcal{L}_{CL_{same}}$:
\begin{equation}
    {cons_{same}}_i = -\log {\frac{e^{sim(z_i,z_i^{drop})/\tau}}{\sum_{j=1}^N e^{sim(z_i,z_j^{drop})/\tau}}}
\end{equation}
\begin{equation}
   \mathcal{L}_{CL_{same}}=\sum_i {cons_{same}}_i 
\end{equation}
where $z_i$ and $z_i^{drop}$ denote the representations of $x_i$ and $x_i^{drop}$, respectively.

At the time of optimization, we take the data pairs $(x_i,x_i^{like})$ and $(x_i,x_i^{drop})$ as 
positives, and other in-batch instances as negatives. By summing them up, the total loss for the contrastive paradigm is: 
\begin{equation}
    \mathcal{L}_{CL} = \mathcal{L}_{CL_{diff}} + \mathcal{L}_{CL_{same}}
\end{equation}

The architecture of our contrastive learning is shown in Fig.\ref{fig:attack}, the Content Retriever retrieves $x_i^{like}$ and $x_i^{drop}$ as positive samples $x_c^+$. 
Finally, incorporating contrasitve learning into our auto-encoder training, the loss of auto-encoder is formalized as follows: 
\begin{equation}
    \mathcal{L}(\theta_{enc},\theta_{dec}) = \mathcal{L}_{rec} + \lambda \mathcal{L}_{CL}
\end{equation}
where $\lambda$ is a balancing hyperparameter.

\subsection{Siamese-Structure Classifier}

% TODO: Add more intuition
When conducting style transfer through gradient-guided update methods, 
the general line is to train a classifier which directly outputs the probability with respect to each selection \cite{cvpr2017plug,19nips,20aaai}. 
However, \cite{19nips} demonstrates that a gradient-guided optimization for text style classifier can become an attack to classifier, where the style of embedding is misclassified. 
Inspired by the superiority of Siamese Networks in various recent models for unsupervised visual representation learning \cite{caron2020unsupervised, chen2021exploring}, 
we adapt a siamese-structure method which decides the style of an embedding by conducting comparison between other label-known samples. 
Provided that our proposed classifier is a comparison based one, the  classifier achieves higher accuracy as the number of its compared samples increases. 
Experiments indicate our proposed classifier structure effectively alleviates the issue of misclassification in the embedding style classifier. 

Inspired by \cite{chen2021exploring}, the siamese-structure based classifier consists of a Style Extractor $e = f(z;\theta_{f})$ and a Style Predictor $r = h(e;\theta_{h})$, which takes as input the output of sentence Encoder and Style Extractor.

To conduct comparison for determining the style of sentence $x_1$ on the basis of label-known sentence $x_2$, they are first fed into the Encoder $Enc$ and Style Extractor $f$ to get corresponding style representations $e_1$ and $e_2$.
Finally, the similarity of known sentence $x_1$ to known sentence $x_2$ is calculated as:
\begin{equation}
    sim(e_1,e_2) = cos(\frac{h(e_1)}{||h(e_1)||}, \frac{e_2}{||e_2||})
\end{equation}
where higher similarity score $sim(e_1,e_2)$ denotes two input sentences $x_1$ and $x_2$ are more likely to be of the same style, lower score denotes they tend to differ in terms of style. 

In order to make full use of labelled data and to ensure diversity of comparison, for $x_i$ with style $s_i$, we randomly sample $n$ 
positive sentences ${x_i}^+$ of the same style to $s_i$ and randomly sample $m$ negative sentences ${x_i}^-$ of different style to $s_i$. 
Ensuring the diversity of the comparisons and the robustness of our siamese-structure, we random sample sentences from positive and negative corpus. 
In such manner, we get data pairs $(x_i,{x_i^0}^+,...,{x_i^n}^+, {x_i^0}^-,...,{x_i^m}^-)$.

% To predict the style of sentence $x_i$, we optimize the loss of siamese-structure classifier as follows: 
In the training phase of siamese-structure based classifier, we optimize: 
\begin{equation}
  l_i^k = -\log {\frac{e^{sim(e_i,{e_i^k}^+)/\tau}}{e^{sim(e_i,{e_i^k}^+)/\tau} + \sum_{j=1}^m e^{sim(e_i,{e_i^j}^-)/\tau}}} \label{eq:2}  \\
\end{equation}
\begin{equation}
  \mathcal{L}_{Sia} = \sum_i l_i = \sum_i \sum_{k=1}^n l_i^k \label{eq:sia}
\end{equation}
where $e$ denotes the representation after feeding $z$, the output embedding of sentence Encoder $Enc$, into the Style Extractor $f$, 
$\mathcal{L}_{Sia}$ denotes the loss function of the siamese structure classifier for optimizing.
It is worth noting that gradient is only back propagated through the Style Predictor side of label-unknown sentences, not through the side of label-known sentences at training time.

The training process of the siamese architecture is shown in the top plot of Fig.\ref{fig:attack}, the Style Retriever retrieves ${x_i^k}^+$ as positive samples $x_s^+$, and ${x_i^k}^-$ as negative samples $x_s^-$. 
We set hyperparameters $n$ and $m$ the same in the training phase and inference phase to avoid the introduction of other prior information.

% Inspired by the superiority of Siamese networks in various recent models for unsupervised visual 
% representation learning siamese representation learning\cite{caron2020unsupervised, chen2021exploring}.
% Hence, we propose a Siamese Structure Classifier, which is based on a comparison between the original 
% sentence and the positive and negative examples to classify. In our framework, we refer to the structure
% of SiaSiam\cite{chen2021exploring} to measure the similarity between latent representations.
% Suppose $z_1$ and $z_2$ are representations of two sentences, $p_1$ and $p_2$ 
% are the respective projection of $z_1$ and $z_2$ where $p=h(z)$ and $h(\cdot)$ is a projection MLP head,
% the similarity function is:
% $$sim(z_1,z_2)=cos(\frac{p_1}{||p_1||}, \frac{stopgrad(z_2)}{||z_2||}),$$
% where stopgrad means stopping the gradient propagation of the tensor in the computational graph.

\subsection{Text Style Transfer}

At the inference stage, the latent embeddings are edited according to the gradient of siamese-structure classifier
and then decode this to the target sentence with desired style.
Given the original sentence $x$ with style $s_{src}$, the inference process of transferring 
to style $s_{tgt}$ is based on the gradient update of continuous latent space.
We first sample $n$ sentences with target style $s_{tgt}$ as positive samples, denoting as $x^{k+}$ where $k$ from $1$ to $n$. 
Similarly, we sample $m$ sentences from styles except for $s_{tgt}$, denoting as $x^{k-}$ where $k$ range from $1$ to $m$. 

Unlike the training stage, a direct gradient for the latent representations is more appropriate for embeddings editing. 
Therefore we adopt a direct loss function for embeddings at the transferring stage, which is: 
\begin{equation}
\begin{aligned}
    \mathcal{L}_{bp} &= - \sum_{i=1}^n sim(f(z),f(z^{i+})) + \sum_{i=1}^{m} sim(f(z),f(z^{i-})) \\ &= - \sum_{i=1}^n sim(e,e^{i+}) + \sum_{i=1}^{m} sim(e,e^{i-})
\end{aligned}
\end{equation}

The representation $z$ of $x$ is edited as follows: 
\begin{equation}
   \hat z=z-Opt(\nabla_z\mathcal{L}_{bp};\theta_{opt})
\end{equation}
where $Opt$ denotes optimizers for applying gradient optimization to original latent embeddings and $\theta_{opt}$ denotes parameters of the optimizer. 
After editing the embeddings in the direction of their gradient, the transferred sentences are generated from the decoder of 
auto-encoder. 
The transfer steps are shown in the bottom plot of Fig.\ref{fig:attack}.
In our experiment, the Adam optimizer \cite{kingma2014adam} is chosen to optimize latent representations.

\section{Experiments}

\subsection{Datasets}

%\noindent\textbf{Data}.
We use two datasets. (1) \textbf{Yelp} dataset, produced by \cite{delete}, contains restaurant reviews with 
positive and negative sentiments. (2) \textbf{Amazon} dataset, produced by \cite{amazon_dataset}, contains 
product reviews on Amazon with positive and negative sentiments.
These two datasets are both commonly-used datasets in text style transfer.
It is worth noting that human-written references are only available in 
test sets and only non-paralleled data is available during training.
The dataset statistics are shown in Table.\ref{table:statistics}.

\begin{table}[htb]
\small
\centering
\begin{tabular}{lllll}
\toprule
Dataset & Style & Train & Dev & Test \\
\hline
\multirow{2}{*}{Yelp} & Positive & 266041 & 2000 & 500 \\ \cline{2-5} 
 & Negative & 177218 & 2000 & 500 \\ \hline
\multirow{2}{*}{Amazon} & Positive & 277228 & 1015 & 500 \\ \cline{2-5} 
 & Negative & 277769 & 985 & 500 \\ \bottomrule
\end{tabular}
\caption{Data Statistics for Yelp and Amazon Dataset}
\label{table:statistics}
\end{table}

\begin{table*}[htbp]
    \small
    \centering
    %\resizebox{.95\columnwidth}{!}{
\begin{tabular}{lcccccc}
\hline
\multicolumn{1}{c}{\multirow{2}{*}{method}} & \multicolumn{6}{c}{Yelp} \\ \cline{2-7} 
\multicolumn{1}{c}{} & Acc$\uparrow$ & PPL$\downarrow$ & human-BLEU$\uparrow$ & self-BLEU$\uparrow$ & human-WMD$\downarrow$ & self-WMD$\downarrow$ \\ \hline
CrossAlign \cite{crossalign} & 74.7 & 71.6 & 6.79 & 20.74 & 0.449 & 0.307 \\ \hline
StyleEmb \cite{multi_decoders} & 17.9 & 76.0 & 16.65 & \textbf{67.43} & \textbf{0.374} & \textbf{0.128} \\ \hline
MultiDec \cite{multi_decoders} & 53.7 & 95.1 & 11.24 & 40.07 & 0.421 & 0.261 \\ \hline
RuleBase \cite{delete} & 83.7 & 85.7 & \textbf{18.02} & 57.36 & 0.376 & 0.260 \\ \hline
DelRetrGen \cite{delete} & 85.0 & 71.7 & 12.62 & 36.75 & 0.393 & 0.278 \\ \hline
ContiSpace \cite{20aaai} & 85.9 & \textbf{47.0} & 8.15 & 18.64 & 0.423 & 0.310 \\ \hline
GBT \cite{19nips} & 88.2 & 130.2 & 9.61 & 29.14 & 0.421 & 0.280 \\ \hline
%Ours(contrastive) & 90.9 & 114.6 & 9.26 & 24.94 & 0.406 & 0.277 \\ \hline
%Ours(siamese) & 89.3 & 122.5 & 9.77 & 28.87 & 0.407 & 0.258 \\ \hline
$\textit{OURS}_{\textit{C+S}}$ & \textbf{91.0} & 100.8 & 12.21 & 34.45 & 0.387 & 0.236 \\ \hline
\multicolumn{1}{c}{\multirow{2}{*}{method}} & \multicolumn{6}{c}{Amazon} \\ \cline{2-7} 
\multicolumn{1}{c}{} & Acc$\uparrow$ & PPL$\downarrow$ & human-BLEU$\uparrow$ & self-BLEU$\uparrow$ & human-WMD$\downarrow$ & self-WMD$\downarrow$ \\ \hline
CrossAlign \cite{crossalign} & 78.6 & \textbf{22.0} & 1.57 & 2.49 & 0.743 & 0.614 \\ \hline
StyleEmb \cite{multi_decoders} & 45.2 & 85.9 & 13.41 & 31.23 & 0.629 & 0.434 \\ \hline
MultiDec \cite{multi_decoders} & 70.8 & 72.0 & 7.87 & 18.24 & 0.685 & 0.527 \\ \hline
RuleBase \cite{delete} & 67.4 & 130.3 & \textbf{31.75} & \textbf{67.75} & 0.483 & 0.233 \\ \hline
DelRetrGen \cite{delete} & 45.7 & 80.3 & 27.14 & 56.44 & \textbf{0.456} & \textbf{0.200} \\ \hline
ContiSpace \cite{20aaai} & 82.7 & 38.7 & 12.87 & 21.88 & 0.598 & 0.419 \\ \hline
GBT \cite{19nips} & 81.0 & 398.8 & 9.56 & 20.1 & 0.660 & 0.495 \\ \hline
%Ours(contrastive) & 84.0 & 284.6 & 11.29 & 22.94 & 0.617 & 0.433 \\ \hline
%Ours(siamese) & 82.7 & 300.5 & 9.66 & 20.17 & 0.628 & 0.450 \\ \hline
$\textit{OURS}_{\textit{C+S}}$ & \textbf{87.5} & 251.3 & 9.79 & 20.0 & 0.594 & 0.413 \\ \hline
\end{tabular}
    \caption{Automatic Evaluation results for Yelp and Amazon datasets. The notation $\uparrow$ means the higher the better and $\downarrow$ the lower the better. We bold the best value for each evaluation criterion.}
    \label{table:auto}
\end{table*}

\begin{table*}[ht]
    \centering
    \small
    \begin{tabularx}{\textwidth}{lXX}
    
    \hline
            &\hspace{11em}Style transfer from \textbf{negative} to \textbf{positive} (Yelp)                                                                               \\ \hline
    Source        & always rude in their tone and always have shitty customer service !                      \\
    Reference     & such nice customer service, they listen to anyones concerns and assist them with it      \\ %\hline
    CrossAlign    & always authentic all other and and they are the food !                                   \\
    StyleEmb      & always rude in their sauce very quiet on actually attitude customer !                    \\
    MultiDec      & always nice is their decent and use the job customer service !                           \\
    RuleBase      & always i was very pleased in their tone and always have shitty customer service !        \\
    DelRetrGen            & i always enjoy going in always their kristen and always have shitty customer service ! \\
    ContiSpace          & they have always been friendly and helpful in their customer service department !        \\ 
    GBT          & always good with their always chop and knowledgeable oatmeal come always good customer ! \\ %\hline
    %$\textit{OURS}_{\textit{C}}$ & always amazing in their books and always have wonderful customer service !               \\
    %$\textit{OURS}_{\textit{S}}$& always nice with their tone and always have shitty and wonderful customer service !      \\
    $\textit{OURS}_{\textit{C+S}}$ & always amazing in their tone and always have wonderful customer service !              \\ \hline
            & \hspace{11em}Style transfer from \textbf{positive} to \textbf{negative} (Yelp)                                                                               \\ \hline
    Source        &      they were so helpful , kind , and reasonably priced .         \\
    Reference     &  They should've been more helpful, kind, and reasonably priced.     \\ %\hline
    CrossAlign    &           they were so helpful , kind , and , very dirty .                       \\
    StyleEmb      &       they were so helpful , kind , and reasonably priced .             \\
    MultiDec      &       they were so helpful , kind , and priced very unprofessional .       \\
    RuleBase      &     they were there were \_num\_ pieces    \\
    DelRetrGen            & but the place was very disappointed and they were they were quite good .  \\
    ContiSpace          &    they were so kind , rude , and over priced .    \\ 
    GBT          & they were so helpful , not whatever , but were really half shit . \\ %\hline
    %$\textit{OURS}_{\textit{C}}$& always amazing in their books and always have wonderful customer service !               \\
    %$\textit{OURS}_{\textit{S}}$& always nice with their tone and always have shitty and wonderful customer service !      \\
    $\textit{OURS}_{\textit{C+S}}$ & they were so gross , disgusting , but even beans . \\\hline
    \end{tabularx}
    \caption{Sentence Example}
    \label{table:example}
    \end{table*}

\begin{table}[ht]
    \centering
    \small
    %\resizebox{.95\columnwidth}{!}{
\begin{tabular}{|l|c|c|c||c|c|c|}
\hline
\multicolumn{1}{|c|}{\multirow{2}{*}{method}} & \multicolumn{3}{c||}{Yelp} & \multicolumn{3}{c|}{Amazon}\\\cline{2-7}
\multicolumn{1}{|c|}{} & Acc & Con & Gra & Acc & Con & Gra \\ \hline
CrossAlign & 2.07& 2.31& 2.46& 2.67& 1.72& 3.01 \\ \hline
MultiDec & 1.75& 3.08& 2.91& 1.95& 2.13& 2.68 \\ \hline
RuleBase & 2.61& 2.76& 2.65& 1.51& \textbf{3.55}& 3.04 \\ \hline
DelRetrGen & 2.97& 2.83& 3.16& 2.33& 3.23& 2.88 \\ \hline
ContiSpace & 3.52& 2.98& \textbf{3.41}& 2.83& 2.67& \textbf{3.45} \\ \hline
GBT & 3.37& 3.23& 3.00& 2.57& 2.68& 2.61  \\ \hline
%Ours(contrastive) & 3.56 & 3.47 & 3.16 & 2.88& 3.04& 3.28 \\ \hline
%Ours(siamese) & 3.46 & 3.18 & 3.02 & 2.80& 2.75& 3.12 \\ \hline
$\textit{OURS}_{\textit{C+S}}$ & \textbf{3.85} & \textbf{3.61} & 3.23 & \textbf{3.08}& 2.75& 3.32 \\ \hline
\end{tabular}
    \caption{Human evaluation}
    \label{table:human}
\end{table}
\subsection{Metrics}
\noindent\textbf{Automatic Evaluation}.
Following previous works \cite{lm_as_dis,styins}, we evaluate whether a text style transfer model is successful 
from three aspects, namely style transfer accuracy, content invariance and language fluency. For accuracy, 
we train a fastText classifier \cite{fasttext} to discriminate different styles.
For content invariance, we use BLEU\footnote{https://github.com/moses-smt/mosesdecoder/blob/master/\\scripts/generic/multi-bleu.perl} \cite{bleu} 
and WMD \cite{wmd}.
Additionally, the prefixes self- and human- represent the generated sentences compared to the original ones and to the human-written references, respectively. 
Previous methods focus on computing the BLEU score. \cite{metric} studies 13 different metrics and proves BLEU, WMD and POS-distance are the best three to evaluate content invariance in the 
domain of text style transfer, and thus we import WMD as an extra criterion.
For language fluency, we measure the perplexity of a sentence with a 5-gram language model SRILM \cite{srilm}. 

%The results are shown in Table \ref{table:auto}.

\noindent\textbf{Human Evaluation}.
Many previous works \cite{19nips,siamese_paralle,lee2020stable} have shown automatic evaluation on human references is not accurate enough.
This demonstrates that automatic evaluation is not persuasive enough 
in the task of style transfer.
Therefore, we conduct a human-written evaluation on models outputs. 
Due to the lack of human labor, we access the output sentences of Yelp and Amazon, and then randomly select 100 sentences for each model with each style.
Following \cite{delete}, we invite 3 workers to evaluate in a blind review manner and score the sentences from three aspects: 
target attribute match  (Att), content invariance (Con) and grammaticality (Gra). The score of each aspect range 
from 1 to 5 where 5 denotes the best and 1 denotes the worst. 
%The human evaluation results are shown in Table \ref{table:human}.

%The human evaluation results indicate our model outperforms previous ones and the refinements work. 
%Some generated instances are shown in Table \ref{table:example}. 

\subsection{Baselines}

%\noindent\textbf{Baselines}.
We conduct comprehensive comparison with previous state-of-the-art models, 
including CrossAlign \cite{crossalign}, StyleEmb \cite{multi_decoders}, MultiDec \cite{multi_decoders}, 
RuleBase \cite{delete}, DelRetrGen \cite{delete}, ContiSpace \cite{20aaai} and GBT \cite{19nips}.
We consider three variants of our model:

\begin{itemize}
    \item $\textit{OURS}_{\textit{C}}$: With modeling of content invariance. 
    \item $\textit{OURS}_{\textit{S}}$: With siamese-structure based classifier .
    \item $\textit{OURS}_{\textit{C+S}}$: With both proposed structures.
\end{itemize}
When the siamese-structure based classifier is not adopted from our model, we use a MLP classifier which outputs the probability of each style directly as \cite{19nips}.
When the contrastive modeling of content consistency is not adopted, the auto-encoder is only trained with the reconstruction loss $L_{rec}$.

\subsection{Experimental Settings}

To demonstrate the effectiveness of our proposed method, we use the same auto-encoder structure as \cite{19nips}. Therefore, GBT is our model without the contrasitve paradigm and the siamese-structure based classifier.

We use two-layer Transformer both for encoder and decoder.
The encoder embedding size and the decoder embedding size are both set to 256.
The balancing hyperparameter $\lambda$ for training auto-encoder is 0.3.
The number of positive samples $n$ and the number of negative samples $m$ are both 10.
The temperature hyperparameter $\tau$ is 0.1. 

% embedding size, positive negative numbers n, m, layers , GPU, lr, tau, lambda, beta, ...
% our contrastive ... 
% our siamese ...

\subsection{Experimental Results}

% It should be noted that a good style transfer method should perform well on all metrics we discussed above.
Automatic evaluation results of the two datasets are presented in Table.\ref{table:auto}.
It should be noted that a good style transfer method should perform well on all aforementioned metrics.
StyleEmb in Yelp dataset achieves perfect result in content retention and sentence fluency, however, it is not a successful method as most sentences ($82.1\%$) are not successfully transferred to the target style.
RuleBase and DelRetrGen in Amazon dataset both reflect similar issue, and thus cannot be considered as accomplish the style transfer task successfully.
We see that our model achieves superior performance in transfer accuracy, outperforming all other models by a large margin. Moreover, our model achieves satisfactory performance in content invariance (BLEU and WMD).
Compared to previous gradient-guided models (ContiSpace and GBT), our model outstands in all aspects except for language fluency.
% Our model also exceeds other gradient-guided models (ContiSpace and GBT) in all aspects, if not less fluent (PPL) than ContiSpace.
% For those previous methods which achieve high content retention (BLEU and WMD) but are not good at the attribute accuracy (\eg StyleEmb).
% This means that those methods only make a few changes to the sentences reduce the conversion rate without improving the conversion rate of the style.
% For metrics WMD, the results show that our model is able to get a higher success rate 
% with fewer words moved than most models.
Nevertheless, our model shows a shortcoming in language fluency (PPL).
This can be due to the inappropriate structure of the Transformer for gradient-based text style transfer, as another Transformer-structure method GBT also performs unsound in terms of sentence fluency. 
In addition, our proposed two structures help relieve the issue of sentence perplexity, as considerable improvement is gained compared to the line without them (GBT).

\begin{figure}[ht]
    \centering
    \includegraphics[width=0.9\columnwidth]{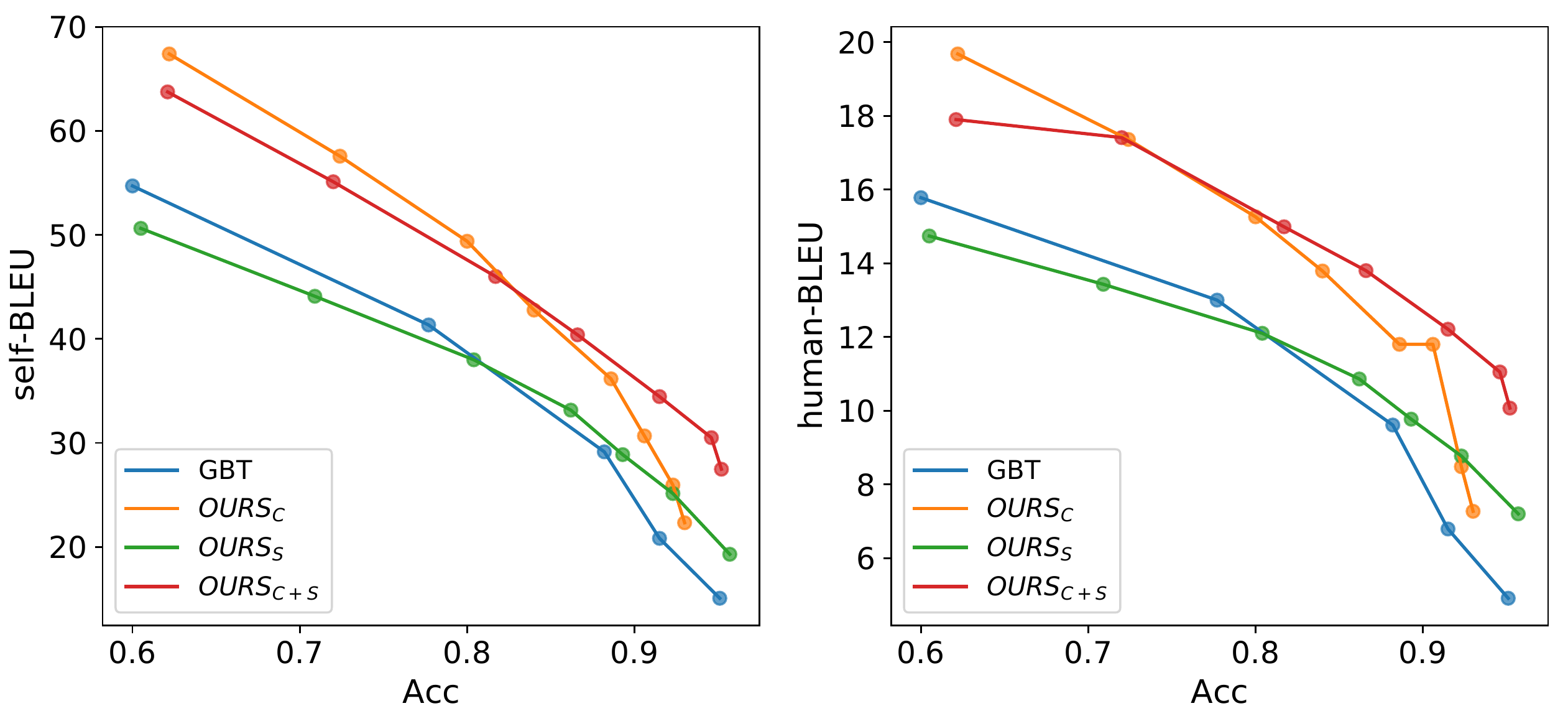}
    \caption{Ablation study for understanding each of the impact of our proposed structure}
    \label{fig:ablation}
\end{figure}

For the human evaluating, we choose five well-performed models according to
the automatic evaluation results as competitors and the results are presented in Table.\ref{table:human}. 
With respect to content invariance (Con) and grammaticality (Gra), we notice some models (\eg RuleBase) performed well on automated metrics, but poorly on the human ones.
As an explanation, these models retain most original words and insert some statements with strong stylistic attributes to the original sentences.
Such line of modifications results in the influence of the original semantics and fluency, which is difficult to be identified by automatic evaluation methods.
Concerning overall human metrics, our model exhibits outstanding performance in the field of style accuracy and content invariance and achieves the best overall performance.
The result of our human evaluation is in accord with the automatic ones.
We compare some generated sentences in Table.\ref{table:example}.

% TODO: 

% \noindent\textbf{Implementaiton}.
% To illustrate that it is our refinements of UNOC and UOEC work instead of other parts of our model, we use the same structure of autoencoder as \cite{19nips}, which is a Transformer-based one \cite{transformer}. In our experiment, we set $n$ to 10 and $m$ to 10.

\subsection{Ablation Study}

To understand the impact of two proposed structures of our proposed model, we further do an ablation study.
We choose to present the results of Yelp dataset in the main text since the results of Amazon dataset reflect similar conclusion.
More details are shown in Appendix. 
The automatic evaluation result is displayed is Table.\ref{table:ablation}.

% on Yelp dataset, and results are shown in Table.\ref{table:ablation}.
% When the siamese-structure based classifier is subtracted from our model, we adopt a conventional MLP classifier which outputs the probability of each style directly as \cite{19nips}.
% When the contrastive modeling of content invariance is removed, the auto-encoder is only trained with the reconstruction loss $L_{rec}$.
% When not using siamese-structure classifier, we adopt a MLP classifier for embeddings which refers to \cite{19nips}.
% When not using contrastive paradigm, we just adopt a standard Transformer auto-encoder framework.

For better understanding the role of the two components, we follow the universal line of focusing
on the relationship between style accuracy and content invariance \cite{delete,styins}, including self-BLEU and human-BLEU. By changing the learning rate and update steps of latent optimizer, 
various data points can be obtained. The results are demonstrated in Fig.\ref{fig:ablation}.

\begin{figure}[t]
    \centering
    \includegraphics[width=0.8\columnwidth]{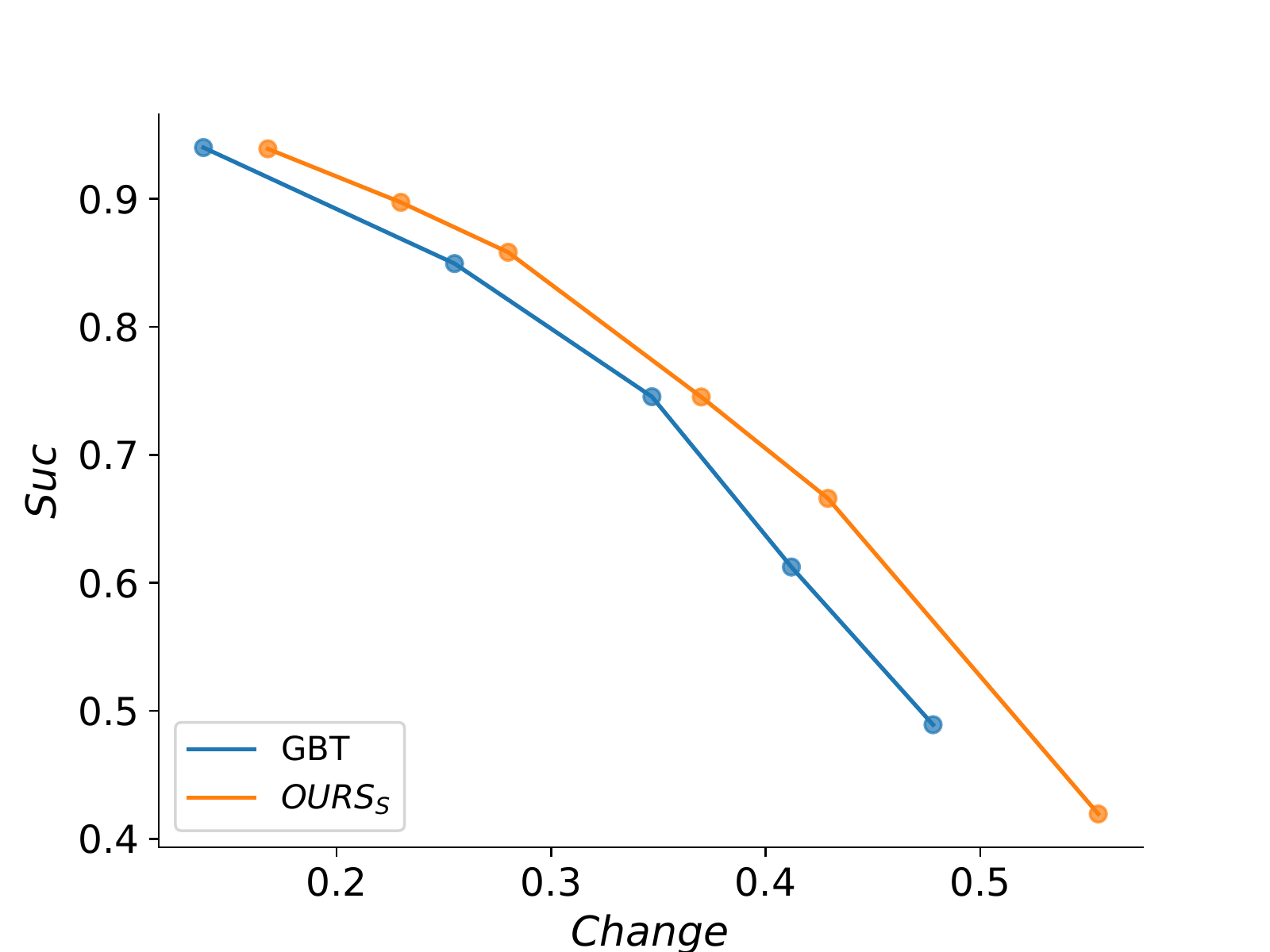}
    \caption{The relationship between $\textit{Change}$ and $\textit{Suc}$.
    % The Change means the numbers of sentence changes in the process of embedding editing.
    % The Suc means the criterion of the robustness of a classifier structure.
    }
    \label{fig:change_suc}
\end{figure}

The two graphs reveal similar conclusions. 
(1) The contrastive paradigm which clusters content similar sentences largely improves overall performance. 
(2) The siamese-structure classifier improves the style accuracy when BLEU is the same, especially when the accuracy rate is high.
We also notice the siamese structure classifier performs slightly worse than the conventional one when style accuracy is low.
This can be due to siamese-structure classifier which make decision by comparing latent representations might introduce other irrelevant content from other compared references. 
This is inevitable as long as the content distribution of two styles 
are not the same \cite{li2020complementary}.
Moreover, the gain of relieving misclassification issue of embedding classifier outstands the loss of irrelevant content export when transfer extent increases considerably.

% \begin{figure}[ht]
%     \centering
%     \includegraphics[width=0.9\columnwidth]{acc_self-bleu}
%     \caption{The relationship between acc and self-bleu}
%     \label{fig:acc_self}
% \end{figure}

% \begin{figure}[ht]
%     \centering
%     \includegraphics[width=0.9\columnwidth]{acc_human-bleu} 
%     \caption{The relationship between acc and human-bleu}
%     \label{fig:acc_human}
% \end{figure}

\subsection{Resistance to Style Misclassification}

When a latent representation alters to be a new one, the predicted label 
of new latent might switch while their decoded sentences are the same, which is a misclassification case that an expected conversion becomes an attack to classifier.
Actually, there are four cases when altering a representation.
%When an embedding updates minimally in the direction of its gradient, 
%there are four possible cases:

\begin{enumerate}
   \item Predicted label unchanged, decoded sentence unchanged.
   \item Predicted label changed, decoded sentence unchanged.
   \item Predicted label unchanged, decoded sentence change.
   \item Predicted label changed, decoded sentence changed.
\end{enumerate}

\begin{table}[t]
    \small
    \resizebox{.95\columnwidth}{!}{
\begin{tabular}{rcccc}
\hline
\multicolumn{1}{c}{\multirow{2}{*}{Model}} & \multicolumn{4}{c}{Yelp} \\ \cline{2-5} 
\multicolumn{1}{c}{} & Acc$\uparrow$ & PPL$\downarrow$ & human-BLEU$\uparrow$ & self-BLEU$\uparrow$ \\ \hline
\multicolumn{1}{l}{$\textit{OURS}_{\textit{C+S}}$} & 91.0 & 100.8 & 12.21 & 34.45 \\ \hline
-S & 90.9 & 114.6 & 9.26 & 24.94 \\ 
-C & 89.3 & 122.5 & 9.77 & 28.87 \\ 
-C-S & 88.2 & 130.2 & 9.61 & 29.14 \\ \hline
\end{tabular}}
\caption{Model ablation study results on Yelp dataset. Letter C denotes contrastive paradigm for training auto-encoder and Letter S denotes siamese-structure classifier.}
\label{table:ablation}
\end{table}

The first condition ($\textit{Keep}$) denotes gradient optimization do not 
influence the judgement of the classifier, and the second condition ($\textit{Attack}$) 
denotes an attack to the classifier.
We evaluate the robustness of a classifier structure by criterion:
\begin{equation}
    \textit{Suc}=\frac{\textit{Keep}}{\textit{Keep}+\textit{Attack}}
\end{equation}
where higher $\textit{Suc}$ means less vulnerable to classifier attack and more resistant to classifier 
misclassification.
However, directly measuring the robustness of a classifier structure by 
$\textit{Suc}$ is unfair, since $\textit{Suc}$ should be close to 1 when the embedding
optimization speed is especially minimal.
Therefore, we measure the impact of optimization speed with the proportion of changed sentences $\textit{Change}$, which equals the sum of the second condition add fourth condition. 
%The relationship 
%between $Change$ and $Suc$ Figure \ref{fig:change_suc} indicates that the siamese 
%structure classifier outperforms the conventional discriminator in all conditions.
Finally, we use a MLP embedding classifier in GBT which directly outputs the probability of each class as a competitor.
Fig.\ref{fig:change_suc} and Fig.\ref{fig:ablation} indicate that our siamese-structure classifier is more resistant to classifier 
misclassification and improves transfer accuracy in the task of text style transfer, respectively.

% \section{Future Work}
% I don't want to work anymore.

\section{Conclusion}
In this work, we propose a novel gradient-guided framework for unsupervised text style transfer, which solves two issues of previous gradient-based works.
We propose a contrastive paradigm for training the auto-encoder to gain better content consistency and design a siamese-structure classifier to alleviate the misclassification issue of embedding classifier.
Our experiments results show that our approach achieves state-of-the-art performance.

%In future work, due to the lack of quality data in this domain, 
%we plan to explore improve the performance with low resources, allowing use
%to get better pseudo-parallel samples in the stage of contrastive learning. 
%Meanwhile, this is more in line with practical style transfer situation.

%\bibliography{aaai}

\bibliography{aaai22.bib}

\begin{thebibliography}{37}
\providecommand{\natexlab}[1]{#1}

\bibitem[{Carlsson et~al.(2020)Carlsson, Gyllensten, Gogoulou, Hellqvist, and
  Sahlgren}]{carlsson2020semantic}
Carlsson, F.; Gyllensten, A.~C.; Gogoulou, E.; Hellqvist, E.~Y.; and Sahlgren,
  M. 2020.
\newblock Semantic re-tuning with contrastive tension.
\newblock In \emph{International Conference on Learning Representations}.

\bibitem[{Caron et~al.(2020)Caron, Misra, Mairal, Goyal, Bojanowski, and
  Joulin}]{caron2020unsupervised}
Caron, M.; Misra, I.; Mairal, J.; Goyal, P.; Bojanowski, P.; and Joulin, A.
  2020.
\newblock Unsupervised learning of visual features by contrasting cluster
  assignments.
\newblock arXiv:2006.09882.

\bibitem[{Chen et~al.(2020)Chen, Kornblith, Swersky, Norouzi, and
  Hinton}]{chen2020big}
Chen, T.; Kornblith, S.; Swersky, K.; Norouzi, M.; and Hinton, G. 2020.
\newblock Big self-supervised models are strong semi-supervised learners.
\newblock arXiv:2006.10029.

\bibitem[{Chen and He(2021)}]{chen2021exploring}
Chen, X.; and He, K. 2021.
\newblock Exploring simple siamese representation learning.
\newblock In \emph{Proceedings of the IEEE/CVF Conference on Computer Vision
  and Pattern Recognition}, 15750--15758.

\bibitem[{Fu et~al.(2018)Fu, Tan, Peng, Zhao, and Yan}]{multi_decoders}
Fu, Z.; Tan, X.; Peng, N.; Zhao, D.; and Yan, R. 2018.
\newblock Style transfer in text: Exploration and evaluation.
\newblock In \emph{Proceedings of the AAAI Conference on Artificial
  Intelligence}, volume~32.

\bibitem[{Gao, Yao, and Chen(2021)}]{gao2021simcse}
Gao, T.; Yao, X.; and Chen, D. 2021.
\newblock SimCSE: Simple Contrastive Learning of Sentence Embeddings.
\newblock arXiv:2104.08821.

\bibitem[{Hadsell, Chopra, and LeCun(2006)}]{hadsell2006dimensionality}
Hadsell, R.; Chopra, S.; and LeCun, Y. 2006.
\newblock Dimensionality reduction by learning an invariant mapping.
\newblock In \emph{2006 IEEE Computer Society Conference on Computer Vision and
  Pattern Recognition (CVPR'06)}, volume~2, 1735--1742. IEEE.

\bibitem[{He et~al.(2020)He, Fan, Wu, Xie, and Girshick}]{he2020momentum}
He, K.; Fan, H.; Wu, Y.; Xie, S.; and Girshick, R. 2020.
\newblock Momentum contrast for unsupervised visual representation learning.
\newblock In \emph{Proceedings of the IEEE/CVF Conference on Computer Vision
  and Pattern Recognition}, 9729--9738.

\bibitem[{He and McAuley(2016)}]{amazon_dataset}
He, R.; and McAuley, J. 2016.
\newblock Ups and downs: Modeling the visual evolution of fashion trends with
  one-class collaborative filtering.
\newblock In \emph{proceedings of the 25th international conference on world
  wide web}, 507--517.

\bibitem[{Hsieh et~al.(2019)Hsieh, Cheng, Juan, Wei, Hsu, and Hsieh}]{19attack}
Hsieh, Y.-L.; Cheng, M.; Juan, D.-C.; Wei, W.; Hsu, W.-L.; and Hsieh, C.-J.
  2019.
\newblock Natural Adversarial Sentence Generation with Gradient-based
  Perturbation.
\newblock arXiv:1909.04495.

\bibitem[{Hu et~al.(2017)Hu, Yang, Liang, Salakhutdinov, and
  Xing}]{towardscontrolled}
Hu, Z.; Yang, Z.; Liang, X.; Salakhutdinov, R.; and Xing, E.~P. 2017.
\newblock Toward Controlled Generation of Text.
\newblock In \emph{ICML}.

\bibitem[{Huang et~al.(2019)Huang, Wu, Wei, and Luan}]{huang2019dictionary}
Huang, S.; Wu, Y.; Wei, F.; and Luan, Z. 2019.
\newblock Dictionary-guided editing networks for paraphrase generation.
\newblock In \emph{Proceedings of the AAAI Conference on Artificial
  Intelligence}, volume~33, 6546--6553.

\bibitem[{Jin et~al.(2020)Jin, Jin, Zhou, Orii, and Szolovits}]{headline1}
Jin, D.; Jin, Z.; Zhou, J.~T.; Orii, L.; and Szolovits, P. 2020.
\newblock Hooks in the Headline: Learning to Generate Headlines with Controlled
  Styles.
\newblock In \emph{Proceedings of the 58th Annual Meeting of the Association
  for Computational Linguistics}, 5082--5093.

\bibitem[{John et~al.(2019)John, Mou, Bahuleyan, and
  Vechtomova}]{john2019disentangled}
John, V.; Mou, L.; Bahuleyan, H.; and Vechtomova, O. 2019.
\newblock Disentangled Representation Learning for Non-Parallel Text Style
  Transfer.
\newblock In \emph{Proceedings of the 57th Annual Meeting of the Association
  for Computational Linguistics}, 424--434.

\bibitem[{Joulin et~al.(2017)Joulin, Grave, Bojanowski, and Mikolov}]{fasttext}
Joulin, A.; Grave, {\'E}.; Bojanowski, P.; and Mikolov, T. 2017.
\newblock Bag of Tricks for Efficient Text Classification.
\newblock In \emph{Proceedings of the 15th Conference of the European Chapter
  of the Association for Computational Linguistics: Volume 2, Short Papers},
  427--431.

\bibitem[{Kaushik, Hovy, and Lipton(2019)}]{kaushik2019learning}
Kaushik, D.; Hovy, E.; and Lipton, Z.~C. 2019.
\newblock Learning the difference that makes a difference with
  counterfactually-augmented data.
\newblock arXiv:1909.12434.

\bibitem[{Khosla et~al.(2020)Khosla, Teterwak, Wang, Sarna, Tian, Isola,
  Maschinot, Liu, and Krishnan}]{supervised_contrastive}
Khosla, P.; Teterwak, P.; Wang, C.; Sarna, A.; Tian, Y.; Isola, P.; Maschinot,
  A.; Liu, C.; and Krishnan, D. 2020.
\newblock Supervised Contrastive Learning.
\newblock \emph{Advances in Neural Information Processing Systems}, 33.

\bibitem[{Kim and Sohn(2020)}]{siamese_paralle}
Kim, H.; and Sohn, K.-A. 2020.
\newblock How Positive Are You: Text Style Transfer using Adaptive Style
  Embedding.
\newblock In \emph{Proceedings of the 28th International Conference on
  Computational Linguistics}, 2115--2125.

\bibitem[{Kingma and Ba(2014)}]{kingma2014adam}
Kingma, D.~P.; and Ba, J. 2014.
\newblock Adam: A method for stochastic optimization.
\newblock arXiv:1412.6980.

\bibitem[{Kusner et~al.(2015)Kusner, Sun, Kolkin, and Weinberger}]{wmd}
Kusner, M.; Sun, Y.; Kolkin, N.; and Weinberger, K. 2015.
\newblock From word embeddings to document distances.
\newblock In \emph{International conference on machine learning}, 957--966.
  PMLR.

\bibitem[{Lee(2020)}]{lee2020stable}
Lee, J. 2020.
\newblock Stable Style Transformer: Delete and Generate Approach with
  Encoder-Decoder for Text Style Transfer.
\newblock In \emph{Proceedings of the 13th International Conference on Natural
  Language Generation}, 195--204.

\bibitem[{Li et~al.(2018)Li, Jia, He, and Liang}]{delete}
Li, J.; Jia, R.; He, H.; and Liang, P. 2018.
\newblock Delete, Retrieve, Generate: a Simple Approach to Sentiment and Style
  Transfer.
\newblock In \emph{Proceedings of the 2018 Conference of the North American
  Chapter of the Association for Computational Linguistics: Human Language
  Technologies, Volume 1 (Long Papers)}, 1865--1874.

\bibitem[{Li et~al.(2021)Li, Chen, Yang, Gao, Zhao, and Yan}]{headline2}
Li, M.; Chen, X.; Yang, M.; Gao, S.; Zhao, D.; and Yan, R. 2021.
\newblock The Style-Content Duality of Attractiveness: Learning to Write
  Eye-Catching Headlines via Disentanglement.
\newblock In \emph{Proceedings of the AAAI Conference on Artificial
  Intelligence}, volume~35, 13252--13260.

\bibitem[{Li et~al.(2020)Li, Li, Zhang, Li, Zheng, Carin, and
  Gao}]{li2020complementary}
Li, Y.; Li, C.; Zhang, Y.; Li, X.; Zheng, G.; Carin, L.; and Gao, J. 2020.
\newblock Complementary auxiliary classifiers for label-conditional text
  generation.
\newblock In \emph{Proceedings of the AAAI Conference on Artificial
  Intelligence}, 05, 8303--8310.

\bibitem[{Liu et~al.(2020)Liu, Fu, Zhang, Pal, and Lv}]{20aaai}
Liu, D.; Fu, J.; Zhang, Y.; Pal, C.; and Lv, J. 2020.
\newblock Revision in continuous space: Unsupervised text style transfer
  without adversarial learning.
\newblock In \emph{Proceedings of the AAAI Conference on Artificial
  Intelligence}, 8376--8383.

\bibitem[{Nguyen et~al.(2017)Nguyen, Clune, Bengio, Dosovitskiy, and
  Yosinski}]{cvpr2017plug}
Nguyen, A.; Clune, J.; Bengio, Y.; Dosovitskiy, A.; and Yosinski, J. 2017.
\newblock Plug \& Play Generative Networks: Conditional Iterative Generation of
  Images in Latent Space.
\newblock In \emph{CVPR}.

\bibitem[{Papineni et~al.(2002)Papineni, Roukos, Ward, and Zhu}]{bleu}
Papineni, K.; Roukos, S.; Ward, T.; and Zhu, W.-J. 2002.
\newblock Bleu: a method for automatic evaluation of machine translation.
\newblock In \emph{Proceedings of the 40th annual meeting of the Association
  for Computational Linguistics}, 311--318.

\bibitem[{Pryzant et~al.(2020)Pryzant, Martinez, Dass, Kurohashi, Jurafsky, and
  Yang}]{pryzant2020automatically}
Pryzant, R.; Martinez, R.~D.; Dass, N.; Kurohashi, S.; Jurafsky, D.; and Yang,
  D. 2020.
\newblock Automatically neutralizing subjective bias in text.
\newblock In \emph{Proceedings of the aaai conference on artificial
  intelligence}. AAAI Press.

\bibitem[{Reimers and Gurevych(2019)}]{reimers2019sentence}
Reimers, N.; and Gurevych, I. 2019.
\newblock Sentence-bert: Sentence embeddings using siamese bert-networks.
\newblock arXiv:1908.10084.

\bibitem[{Shen et~al.(2017)Shen, Lei, Barzilay, and Jaakkola}]{crossalign}
Shen, T.; Lei, T.; Barzilay, R.; and Jaakkola, T. 2017.
\newblock Style transfer from non-parallel text by cross-alignment.
\newblock In \emph{Proceedings of the 31st International Conference on Neural
  Information Processing Systems}, 6833--6844.

\bibitem[{Stolcke(2002)}]{srilm}
Stolcke, A. 2002.
\newblock SRILM-an extensible language modeling toolkit.
\newblock In \emph{Seventh international conference on spoken language
  processing}.

\bibitem[{Tran, Zhang, and Soleymani(2020)}]{tran2020towards}
Tran, M.; Zhang, Y.; and Soleymani, M. 2020.
\newblock Towards A Friendly Online Community: An Unsupervised Style Transfer
  Framework for Profanity Redaction.
\newblock In \emph{Proceedings of the 28th International Conference on
  Computational Linguistics}, 2107--2114.

\bibitem[{Vaswani et~al.(2017)Vaswani, Shazeer, Parmar, Uszkoreit, Jones,
  Gomez, Kaiser, and Polosukhin}]{transformer}
Vaswani, A.; Shazeer, N.; Parmar, N.; Uszkoreit, J.; Jones, L.; Gomez, A.~N.;
  Kaiser, {\L}.; and Polosukhin, I. 2017.
\newblock Attention is all you need.
\newblock In \emph{Advances in neural information processing systems},
  5998--6008.

\bibitem[{Wang, Hua, and Wan(2019)}]{19nips}
Wang, K.; Hua, H.; and Wan, X. 2019.
\newblock Controllable unsupervised text attribute transfer via editing
  entangled latent representation.
\newblock \emph{Advances in Neural Information Processing Systems}, 32:
  11036--11046.

\bibitem[{Yamshchikov et~al.(2021)Yamshchikov, Shibaev, Khlebnikov, and
  Tikhonov}]{metric}
Yamshchikov, I.~P.; Shibaev, V.; Khlebnikov, N.; and Tikhonov, A. 2021.
\newblock Style-transfer and Paraphrase: Looking for a Sensible Semantic
  Similarity Metric.
\newblock In \emph{Proceedings of the AAAI Conference on Artificial
  Intelligence}, 16, 14213--14220.

\bibitem[{Yang et~al.(2018)Yang, Hu, Dyer, Xing, and
  Berg-Kirkpatrick}]{lm_as_dis}
Yang, Z.; Hu, Z.; Dyer, C.; Xing, E.~P.; and Berg-Kirkpatrick, T. 2018.
\newblock Unsupervised text style transfer using language models as
  discriminators.
\newblock In \emph{Proceedings of the 32nd International Conference on Neural
  Information Processing Systems}, 7298--7309.

\bibitem[{Yi et~al.(2020)Yi, Liu, Li, and Sun}]{styins}
Yi, X.; Liu, Z.; Li, W.; and Sun, M. 2020.
\newblock Text Style Transfer via Learning Style Instance Supported Latent
  Space.
\newblock In \emph{IJCAI}, 3801--3807.

\end{thebibliography}

\end{document}